\documentclass[10pt,twocolumn,letterpaper]{article}

\usepackage{cvpr}              

\usepackage{graphicx}
\usepackage{amsmath}
\usepackage{amssymb}
\usepackage{booktabs}

\usepackage[pagebackref,breaklinks,colorlinks]{hyperref}

\usepackage[capitalize]{cleveref}
\crefname{section}{Sec.}{Secs.}
\Crefname{section}{Section}{Sections}
\Crefname{table}{Table}{Tables}
\crefname{table}{Tab.}{Tabs.}

\def\cvprPaperID{7259} 
\def\confName{CVPR}
\def\confYear{2023}

\begin{document}

\title{Deep Seam Prediction for Image Stitching \\ Based on Selection Consistency Loss}

\author{Senmao Cheng ~~~~ Fan Yang ~~~~ Zhi Chen ~~~~ Nanjun Yuan ~~~~ Wenbing Tao\thanks{Corresponding author.}\\
National Key Laboratory of Science and Technology on Multi-spectral Information Processing, \\
School of Artificial Intelligence and Automation, \\
Huazhong University of Science and Technology, Wuhan 430074, China \\
}
\maketitle

\begin{abstract}
	
   Image stitching is to construct panoramic images with wider field of vision (FOV) from some images captured from different viewing positions. To solve the problem of fusion ghosting in the stitched image, seam-driven methods avoid the misalignment area to fuse images by predicting the best seam. Currently, as standard tools of the OpenCV library, dynamic programming (DP) and GraphCut (GC) are still the only commonly used seam prediction methods despite the fact that they were both proposed two decades ago. However, GC can get excellent seam quality but poor real-time performance while DP method has good efficiency but poor seam quality. In this paper, we propose a deep learning based seam prediction method (DSeam) for the sake of high seam quality with high efficiency. To overcome the difficulty of the seam description in network and no  GroundTruth for training we design a selective consistency loss combining the seam shape constraint and seam quality constraint to supervise the network learning. By the constraint of the selection of consistency loss, we implicitly defined the mask boundaries as seams and transform seam prediction into mask prediction. To our knowledge, the proposed DSeam is the first deep learning based seam prediction method for image stitching. Extensive experimental results well demonstrate the superior performance of our proposed Dseam method which is 15 times faster than the classic GC seam prediction method in OpenCV 2.4.9 with similar seam quality. 
\end{abstract}
\vspace{-4mm}
\section{Introduction}
\label{sec:intro}     

Image stitching is a practical and challenging computer vision task which takes multi-views images as input and recovers a panoramic image of them. It can be widely used in many fields including biology \cite{chalfoun2017mist}, medicine \cite{li2017medical}, surveillance video \cite{gaddam2016tiling}, autonomous driving \cite{wang2020multi} and virtual reality \cite{kim2019deep}. Usually, image stitching can be divided into two steps: image registration and image fusion. Image registration estimates the homography parameters of two input images to align them, while image fusion obtain a natural stitched image by fusing the two aligned images. Annoyingly, since the homography transform cannot perfectly align two images with parallax, direct image fusion often result in ghosting in the stitching result which lead to a bad visual experience \cite{burt1983multiresolution,perez2003poisson,nie2021unsupervised}.
\begin{figure}[t]
	\centering
	\includegraphics[width=1\columnwidth]{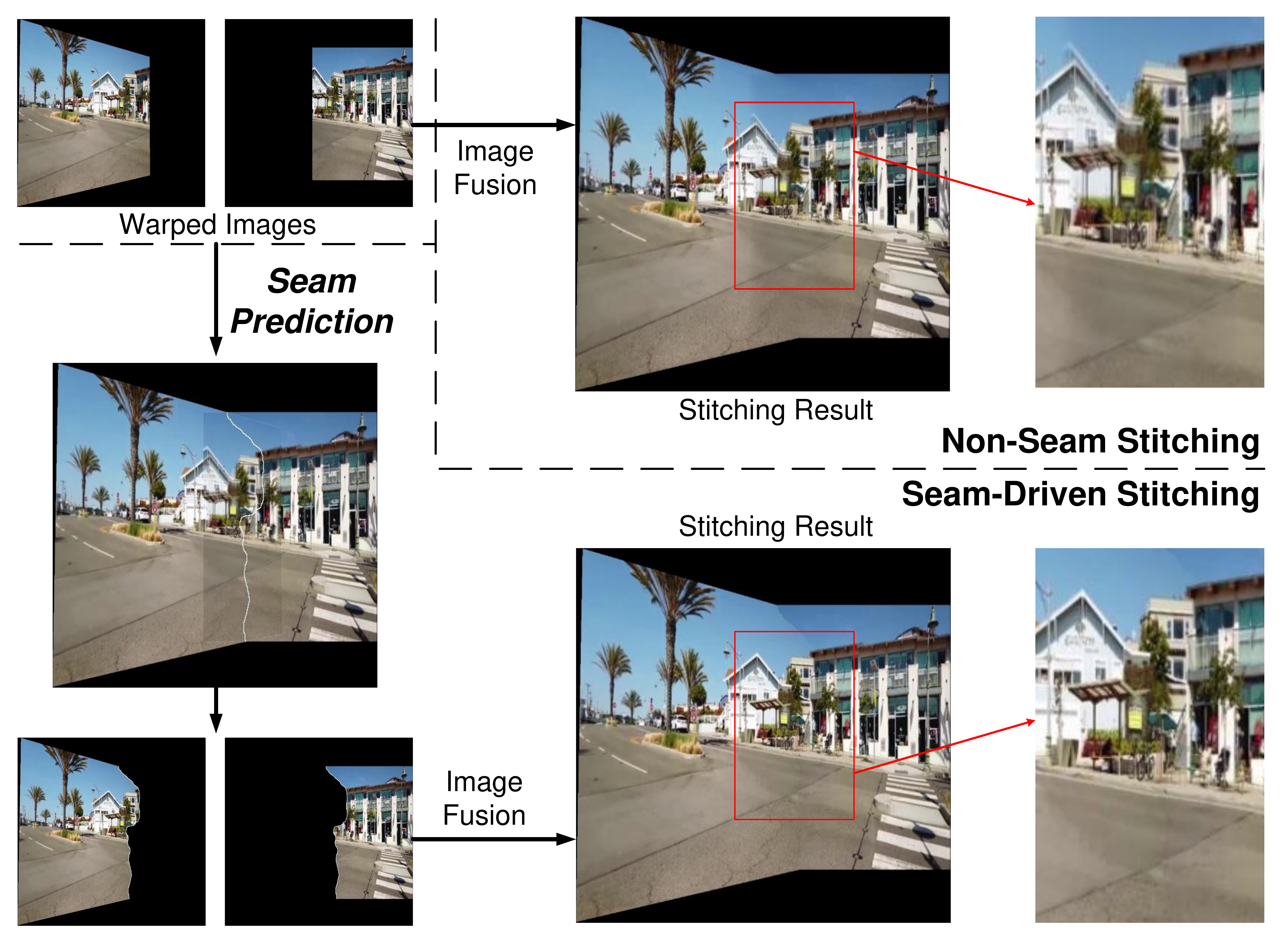} 
	\vspace{-4mm}
	\caption{Non-seam Stitching: Direct image fusion results in severe ghosting. Seam-driven Stitching: Seam prediction and image fusion, the stitching result is quite natural.}
	\vspace{-6mm}
	\label{overlap}
\end{figure}

To eliminate the fusion ghosting, the seam driven stitching method has been proposed, which adds a seam prediction stage before image fusion. As shown in Figure \ref{overlap}, it finds a seam that avoids areas of pixel misalignment to crop the warped images, so that the image fusion can be carried out near the seam without causing fusion ghosting. The key of the seam driven stitching method is to find the best seam in the overlapping area of the two images in order to minimize the impact of parallax. Currently, dynamic programming (DP) \cite{duplaquet1998building} and GraphCut (GC) \cite{kwatra2003graphcut} are still the only two commonly used seam prediction methods and the vast majority of seam driven stitching methods are based on them and their variants despite that they have been proposed for two decades. 
There were several works that try to improve the stitching strategy, such as selecting the one with the optimal seam from multiple homography matrices \cite{gao2013seam}, re-aligning in the area near the seam \cite{zhang2014parallax}, iteratively performing local alignment and seam prediction \cite{lin2016seagull}, etc. However, all these methods are based on the result of seam prediction by 
GC or DP methods and only use seam prediction as a usable component. They do not propose new methods to improve the seam prediction quality or efficiency. 
As standard tools of the OpenCV library \cite{bradski2000opencv}, DP and GC have been widely used for seam prediction in image stitching.
DP\cite{duplaquet1998building} predicts seams by means of path search, which has high speed but poor quality of seams. GC\cite{kwatra2003graphcut} transforms seam prediction into the min-cut problem and iteratively optimizes the energy function to obtain high-quality seams. However, the high computational complexity of the graph cuts optimization in GC-based seam prediction method limits its wide application in the scenarios with real-time requirement. 

In this paper, we try to propose a deep learning based seam prediction method  for the sake of high seam quality with high efficiency.
The most important issue of applying deep learning technique to predict seam is how to define a seam in the network. By definition, the seam is to divide the overlapping area into two parts and avoid the area of pixel misalignment. Thus, the start and end points of the seam must lie on the boundary of the overlapping area, which imposes a very strict constraint on the seam. Meanwhile, the seam requires a high enough degree of freedom to avoid pixel misalignment that may occur anywhere within the overlapping area. Another  challenge needed to be addressed is the lack of GroundTruth data for image stitching. In order to train a network to predict the optimal seam in the overlapping area of the two stitched images, the GroundTruths are usually necessary to supervise the learning of the network. 

To solve the problems mentioned above, we propose a deep seam prediction method (Dseam) based on selection consistency loss combining the seam shape constraint and seam quality constraint. By the constraint of the selection of consistency loss, we transform seam prediction into mask prediction which define a seam implicitly. Specifically, we design an end-to-end network to predict the valid area masks of the input images respectively, and then take the common boundary of the two masks as the seam. We obtain the stitching result based on the predicted masks of the two input images by network and then evaluate the stitching quality according to the selection consistency loss, so as to supervise the training of the network. Therefore, our proposed Dseam method can train a seam prediction network to find the optimal seam in image stitching by an unsupervised way.
Since the boundary dividing the overlapping area is irregular and has a high degree of freedom, the seam defined in this way perfectly fits the above requirements.
In experiments, we evaluate the performance of our method in terms of seam quality and speed. 
Experimental results show that the proposed Dseam has achieved both high
quality and fast speed. It can achieve a speed of 170 FPS with the size of input images being 256×256, which is 15 times faster than the classic GraphCut method in  OpenCV 2.4.9, while having similar seam quality to GraphCut. This greatly benefits the application of our method on real-world scenes.
The contributions of this paper are summarized as follows:

\begin{itemize}
	\item We propose a deep learning based seam prediction method which predict seams in a mask manner. To our knowledge, it is the first time that the deep learning is applied to predict the best seam for image stitching.
	\item We propose the selection consistency loss combining the seam shape constraint and seam quality constraint, which indirectly defines the seam and makes the network trainable by an unsupervised way.
	\item The proposed Dseam method is 15 times faster than the classic GraphCut seam prediction method in OpenCV 2.4.9 with similar seam quality, which is expected to be widely used in practice in the future. 
\end{itemize}

\section{Related Work}

The traditional stitching scheme solves the homography transformation model by image registration \cite{ng2003sift,bay2006surf,rublee2011orb,leutenegger2011brisk,tian2019sosnet,fischler1981random}, and then fuses the warped images obtained by homography transformation to get the stitched image \cite{burt1983multiresolution,perez2003poisson}. To eliminate ghosting in the stitched image, many methods have been proposed. We divide them into non-seam stitching and seam-driven stitching according to whether seam prediction is performed or not.

\subsection{Non-seam Stitching} Since the fusion ghosting is caused by the fact that a single homography matrix cannot perfectly align two images, some methods estimate multiple homography matrices. APAP \cite{zaragoza2013projective} places a mesh on the image and estimates a local homography transformation model for each grid. In order to achieve better alignment, Robust ELA \cite{li2017parallax} combines the grid-based model and the direct deformation strategy. To further preserve image structures in wide-parallax condition, LPC \cite{jia2021leveraging} propose a seam matching strategy leveraging the line-point consistence measure. In low-textured environments, Li et al. \cite{li2015dual} developed a dual-feature warping model for image alignment, using both the sparse feature matches and line correspondences. To overcome the problems of failures for deep learning algorithms in low overlap rate cases, Nie et al. \cite{nie2021depth} design a novel contextual correlation layer (CCL) for multi-grid deep homography estimation. Observing that misalignments in feature-level are more unnoticeable than in pixel-level, UDIS \cite{nie2021unsupervised} fuses images in the feature domain to avoid ghosting. In general, the non-seam stitching method performs well in small-parallax images, but it is hard to deal with large-parallax images which always results in poor stitched images as shown in Figure \ref{UDIS}.

\subsection{Seam Prediction} Seam prediction is the basis of seam-driven stitching method, the traditional seam prediction methods include dynamic programming (DP) \cite{duplaquet1998building} and GraphCut (GC) \cite{kwatra2003graphcut} which have been widely used as standard tools of OpenCV library \cite{bradski2000opencv}. Based on the difference of image color, DP builds a cost map for the overlapping area of the images, and then searches the local optimal path in this map by dynamic programming method. To get the global optimum, GC transforms the seam prediction into a classical min-cut problem and iteratively optimizes the energy function to predict the best seam. Based on the work of GC, Liao and Chen \cite{liao2019quality} proposed a new cost function to calculate the cost map of GC and estimate the seam iteratively, but it may fail at large parallax. For more than a decade, no new seam prediction method has been proposed, DP and GC are still the most commonly used methods for seam prediction in image stitching. 
\begin{figure*}[t]
	\centering
	\includegraphics[width=1\textwidth]{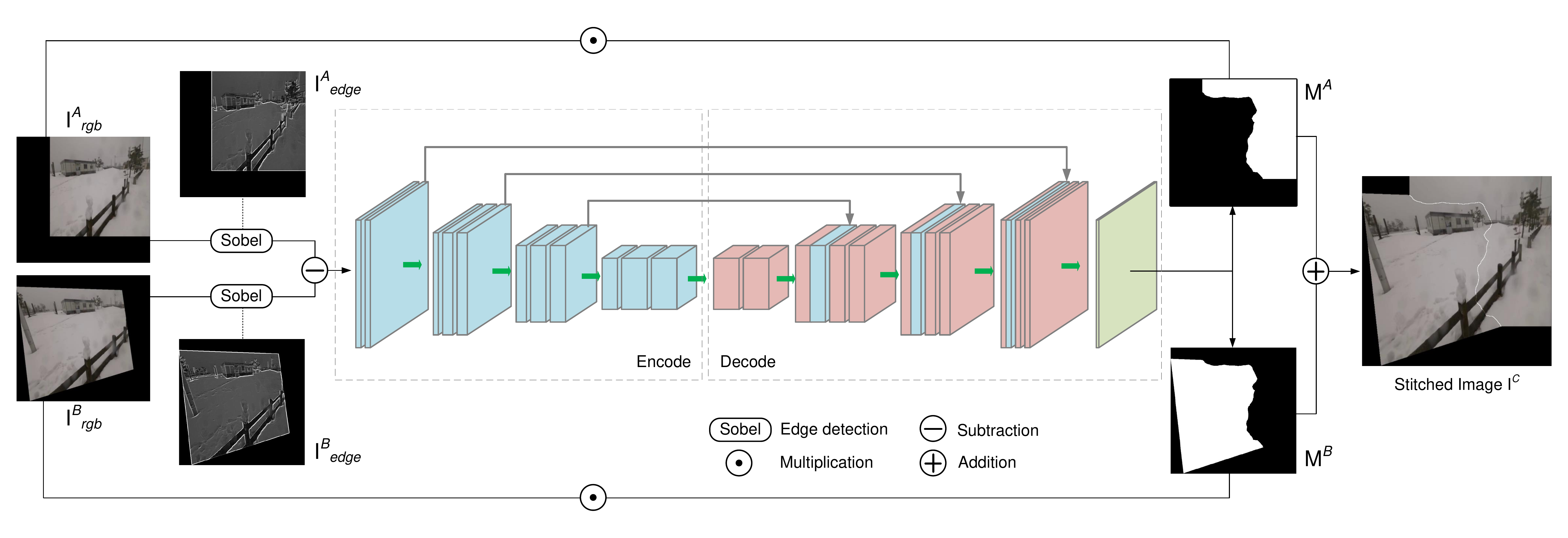} 
	\vspace{-7mm}
	\caption{An overview of our deep seam prediction. }
	\vspace{-5mm}
	\label{network}
\end{figure*}
\subsection{Seam-driven Stitching} Seam-driven stitching predicts a seam to avoid the misalignment area, so that image fusion near the seam will not cause ghosting in the stitched image. Based on the result of seam prediction, many strategies have been tried to achieve better results. In contrast to stressing the global alignment quality, Gao et al. \cite{gao2013seam} propose to choose the best homography with the lowest seam-related cost from candidate homography matrices. Zhang and Liu \cite{zhang2014parallax} propose a local alignment method based on seams, which uses optimal homography to preserve the global image structure. Through iterative warping and seam estimation, Seagull \cite{lin2016seagull} finds the best local stitching area so that the curve and line structure can be protected in the image stitching process. To make full use of the color information, ACIS \cite{li2022automatic} proposes a quaternion rank-1 alignment (QR1A) model to simultaneously learn the optimal seamline and local alignment. Specifically, the seam-driven stitching method only makes some strategy optimization based on the seam predicted by DP or GC, but does not improve the seam prediction method itself. By adopting DP for seam prediction, poor quality seam may lead to a bad stitching result, while GC is too complex to compute in real time.

\section{Methodology}
Image stitching is to project two images with overlapping area onto a common plane to get the warped images \(I^A\in \mathbb{R}^{H\times W}\) and \(I^B\in \mathbb{R}^{H\times W}\), and then fuse them to get a stitched image \(I^C\in \mathbb{R}^{H\times W}\). Due to the existence of parallax, the warped images cannot be perfectly aligned in the overlapping area, which leads to the fusion ghosting in stitched image. To eliminate ghosting, we predict a seam to divide the overlapping area into two parts, each of which takes pixels from only one image. So there will be no fusion ghosting within these two parts of areas. Since pixel misalignment near the seam can also affect the quality of stitched image, we expect the seam to be located in the most similar area of the two images. To get the best seam, we propose the deep seam prediction network.

\subsection{Deep Seam Prediction Network}
\subsubsection{Seam Defined By Mask}
To define a seam for image stitching accurately, it has the following requirements: (1). The seam is a continuous curve. (2). The start and end points of the seam are fixed by the boundary intersection of the input images. (3). It has a high degree of freedom in the overlapping area and can represent arbitrary shapes.

A possible way to meet the high-degree freedom is splines. However, the degree of freedom of splines depends on the number of parameters, and the shape of seam may be very irregular, which is difficult to be represented by a function with finite parameters. So it is hard to directly define a seam by splines. Considering that the seam serves to divide two areas and is the common boundary of them, we define it indirectly by the mask boundary. Thus, we transform the seam prediction into mask prediction, and the seam is determined by the common boundary of two masks. Mask boundary is definitely continuous and has a high enough degree of freedom to form arbitrary shapes. Therefore, we predict the masks \(M^A\in \mathbb{R}^{H\times W}\) and \(M^B\in \mathbb{R}^{H\times W}\) of \(I^A\) and \(I^B\) through the network, and stitched image \(I^C\) can be obtained by the following equation:

\begin{equation}
	I^C=I^A \odot M^A+I^B\odot M^B,
\end{equation}
where \(\odot\) is the pixel-wise multiplication. \(M^A\in \left\lbrace 0,1\right\rbrace\) and \(M^B\in \left\lbrace 0,1\right\rbrace\).

\subsubsection{Network Structure}
After defining the seams as the common boundaries of masks, we design an end-to-end network for mask prediction. As shown in Figure \ref{network}, we take the two warped images as the input of the network, output the masks (\(M^A\) and \(M^B\)) of these two images, and then obtain the stitched image through Equation (1).

Compared with RGB images, edge images only contain the outline of object. After filtering out some invalid information, it may be easier to predict seams from edge images than RGB images. Inspired by this observation, we firstly use the edge extraction operator Sobel \cite{kanopoulos1988design} to transform the input RGB images (\(I^A_{rgb}\) and \(I^B_{rgb}\)) into edge images (\(I^A_{edge}\) and \(I^B_{edge}\)), and then predict a seam from the edge images. Since the loss of seam mainly depends on the difference of pixel values of images, we input the difference image obtained by \(I^A_{edge}-I^B_{edge}\) into the subsequent network.

Our goal is to predict the masks of two images, which is essentially a binary classification problem. For each pixel in the stitched image, the network needs to decide which image it comes from. To obtain better context awareness, we choose the encoding-decoding structure with skip connection between the same resolution \cite{ronneberger2015u} for the backbone of the network. Specifically, the encoding-decoding network is consist of three pooling layers and three deconvolution layers. In addition, skip connections are used to connect low-level and high-level features with the same resolution.

The network finally outputs \(M^{*A}\) and \(M^{*B}\) for image \(I^A\) and \(I^B\), which conform to the equation \(M^{*A} + M^{*B} = \textbf{1}^{H\times W}\) (\(\textbf{1}^{H\times W}\) is an all-one matrix of \(H\times W\)). Since only the areas with valid content need to be supervised, we obtain \(M^{A}\) and \(M^{B}\) shown in Figure \ref{network} by the following equation. 

\begin{equation}
	\begin{aligned}
		M^{A}=M^{*A} \odot M^{AC},\quad
		M^{B}=M^{*B} \odot M^{BC},
	\end{aligned}
\end{equation}
where \(M^{AC}\) and \(M^{BC}\) denote the areas of \(I^A\) and \(I^B\) with valid content (excluding black background areas).

\subsection{Selection Consistency Loss}

In order to make the network output satisfactory masks, we create a new type of loss function, called selection consistency loss. It is designed by the combination of two parts: shape constraint and quality constraint. The former imposes some basic constraints on the seam to satisfy the definition of seam. The latter effectively reflects the quality of seam, which trains the network to predict the best seam.
\begin{figure}[t]
	\centering
	\includegraphics[width=1\columnwidth]{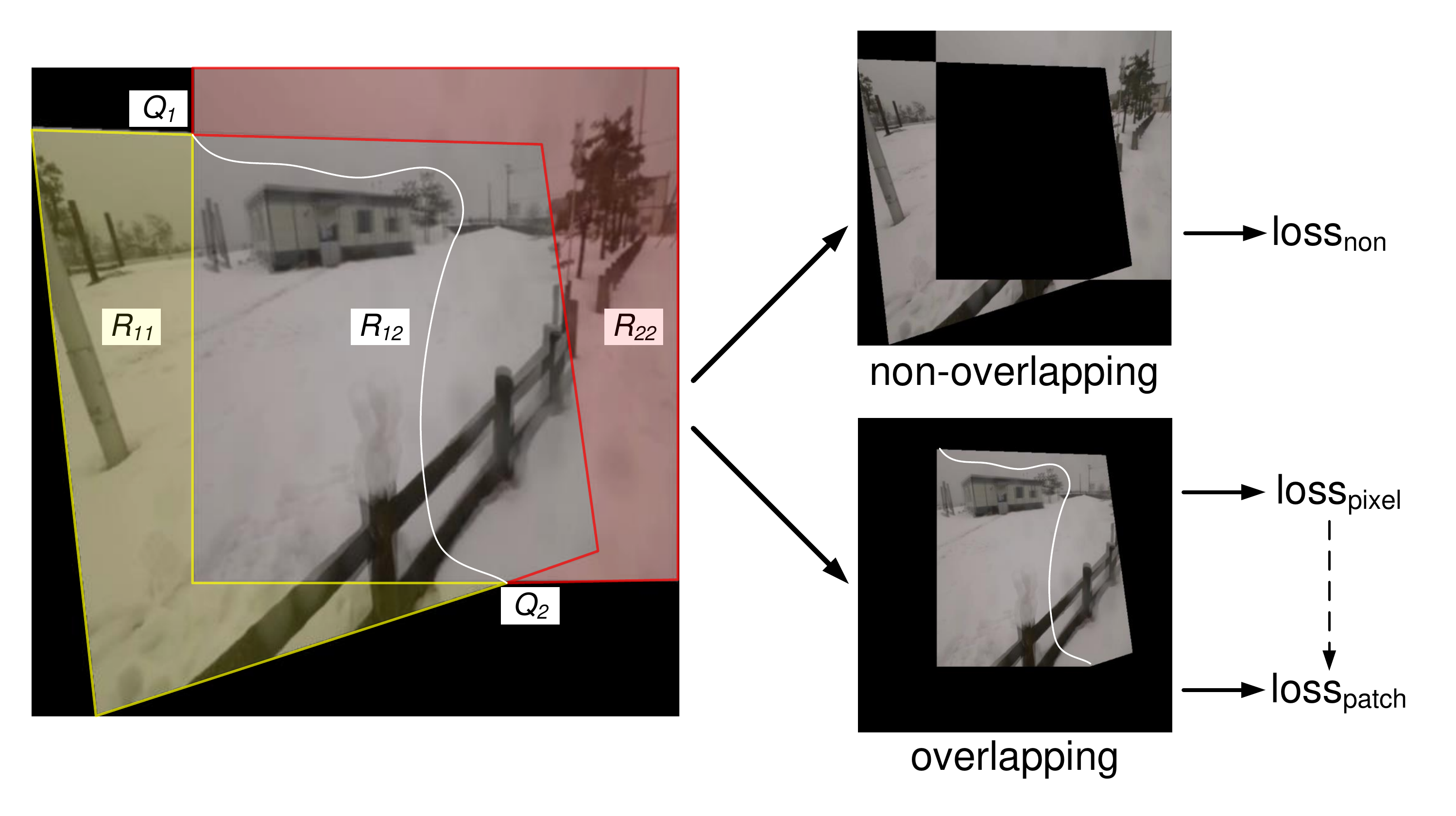} 
	\vspace{-9mm}
	\caption{The two images on the common plane.}
	\vspace{-5mm}
	\label{area}
\end{figure}

\subsubsection{Shape Constraint} 

Since the final performance of the seam is represented by the stitched image \(I^C\), to find a high-quality seam, we need to make some constraints on \(I^C\). As shown in Figure \ref{area}, images \(I^A\) and \(I^B\) are projected on a common plane after registration. The whole plane can be divided into three parts: \(\mathcal{R}_{11}\), \(\mathcal{R}_{22}\) and \(\mathcal{R}_{12}\), which respectively represent the non-overlapping area of \(I^A\),  the non-overlapping area of \(I^B\) and the overlapping area between \(I^A\) and \(I^B\). Since the seam is the dividing line of the two images, for ideal stitching result we expect 
\(I^C(p_i)=I^A(p_i), {p_i\in\mathcal{R}_{11}}\)
and
\(I^C(p_i)=I^B(p_i), {p_i\in\mathcal{R}_{22}}\)
, from which we can derive the loss of the non-overlapping area:
\begin{equation}
	\begin{aligned}
		loss_{non}= &\frac{1}{N_{11}}\sum_{p_i\in \mathcal{R}_{11}}|I^C(p_i)-I^A(p_i)|\\
		+&\frac{1}{N_{22}}\sum_{p_i\in \mathcal{R}_{22}}|I^C(p_i)-I^B(p_i)|,
	\end{aligned}
\end{equation}
where \(I^C(p_i)\), \(I^A(p_i)\) and \(I^B(p_i)\) denote the values of pixel \(p_i\) in the images \(I^C\), \(I^A\) and \(I^B\) respectively.  \(N_{11}\) and \(N_{22}\) are respectively the pixel numbers in \(\mathcal{R}_{11}\) and \(\mathcal{R}_{22}\).

As shown in Figure \ref{area}, \(Q_1\) and \(Q_2\) are fixed demarcation points of the mask of \(I^A\) and \(I^B\). From the definition of the seam, as the seam is the dividing line of the valid area, it must pass through \(Q_1\) and \(Q_2\). Meanwhile, the two points are located on the boundary of the mask of \(I^A\) and \(I^B\). Therefore, \(Q_1\) and \(Q_2\) must be the start and end point of the seam respectively. After \(\mathcal{R}_{11}\) and \(\mathcal{R}_{22}\) are determined by \(loss_{non}\), \(Q_1\) and \(Q_2\) as the intersection of these two areas are naturally determined. Therefore, \(loss_{non}\) can constrain the start and end points of seams.

After the start and end points are determined, we need to further constrain the seam in the \(\mathcal{R}_{12}\). Most fusion methods expect \(I^C(p_i)=(I^A(p_i)+I^B(p_i))/2, {p_i\in\mathcal{R}_{12}}\)
, so that the image \(I^C\) is consistent with \(I^A\) and \(I^B\) in the overlapping area. But this strategy will make each pixel of \(I^C\) be affected by \(I^A\) and \(I^B\) at the same time, forming a fusion ghosting. To address this issue, we create the selection consistency loss of pixel. We expect that each pixel \({p_i}\) of \(I^C\) in the overlapping area comes from only one of \(I^A\) and \(I^B\), which means \(I^C(p_i)=I^A(p_i)\) or \(I^C(p_i)=I^B(p_i)\) for each \({p_i\in\mathcal{R}_{12}}\). As for whether the pixel is consistent with \(I^A\) or \(I^B\), we leave it to the network to select. So we can get the selection consistency loss of the overlapped area:
\begin{equation}
	\begin{aligned}
		loss_{pixel}=\frac{1}{N_{12}}\sum_{p_i\in \mathcal{R}_{12}}min(&|I^C(p_i)-I^A(p_i)|,\\&
		|I^C(p_i)-I^B(p_i)|).
	\end{aligned}
\end{equation}
where \(N_{12}\) is the pixel number of \(\mathcal{R}_{12}\).
According to the definition of seam, \(loss_{pixel}\) only imposes minimal constraints on image \(I^C\) in the overlapping area. The seam defined in this way have the highest degree of freedom and can form arbitrary shapes.

\subsubsection{Quality Constraint} 

Although \(loss_{non}\) and \(loss_{pixel}\) define a seam that satisfies the shape requirements, they do not indicate the location of the seam except for the start and end points. Instead, they only transform the mask boundary into a reasonable seam. In fact, the location determines the seam quality and the seam should be located in the most similar area of the two images. To further meet this requirement, we introduce the quality constraint into the proposed selection consistency loss to avoid pixel misalignment areas.
\begin{figure}[h]
	\centering
	\vspace{-3mm}
	\includegraphics[width=0.9\columnwidth]{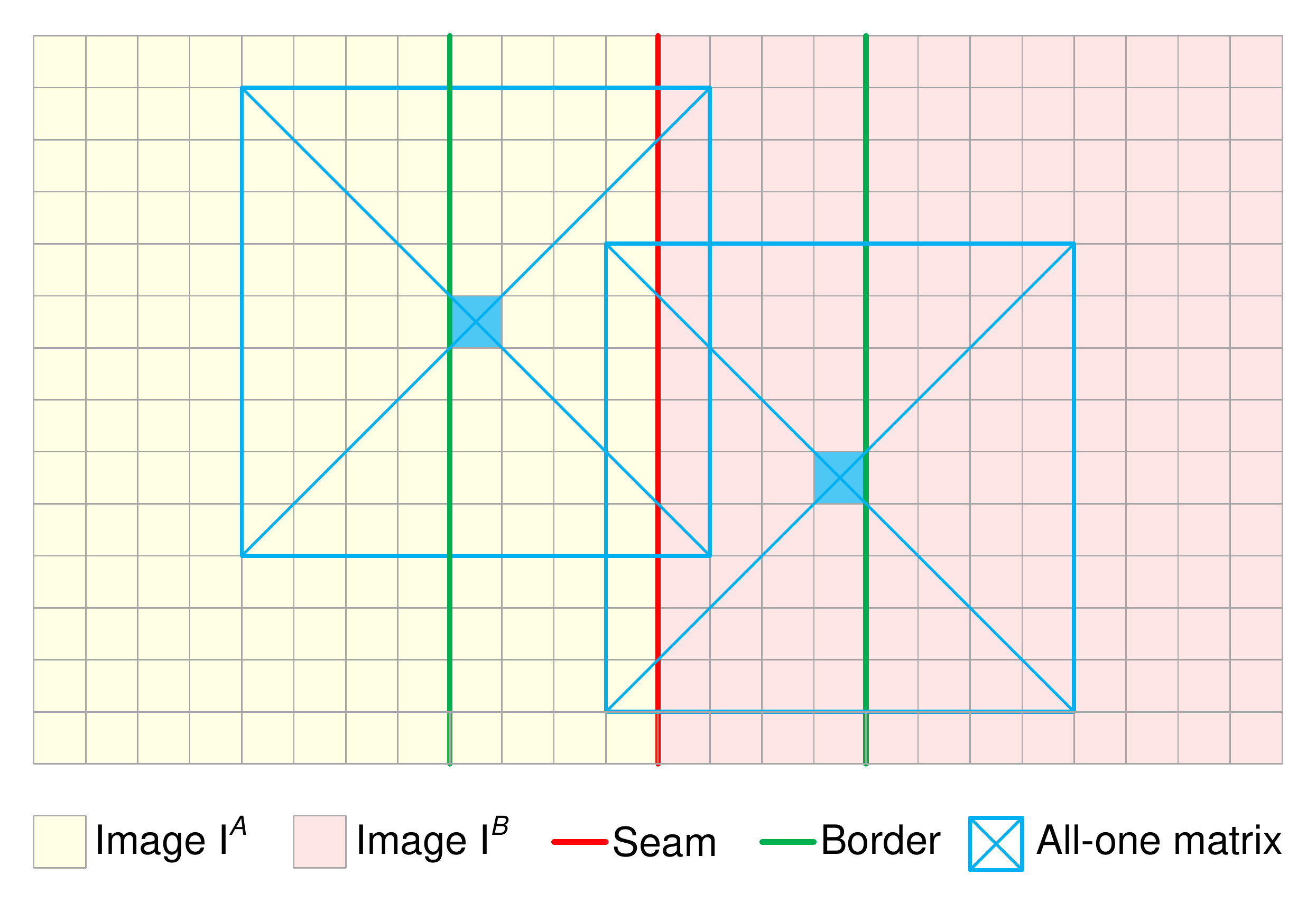} 
	\vspace{-3mm}
	\caption{The area near the seam of \(I^C\). The all-one matrix is used to average filter the \(I^C\) and its size (M) is set to 9 in this paper.}
	\vspace{-3mm}
	\label{matrix}
\end{figure}

Considering that the quality of seam depends on the similarity of images \(I^A\) and \(I^B\) in the vicinity of the seam, we select the area near the seam for analysis. As shown in Figure \ref{matrix}, we take the seam as the center to delimit an image band (M-1) pixels wide as the seam area, and its boundaries are marked with green lines. We expect the pixel difference between \(I^A\) and \(I^B\) in the seam area as loss. To achieve it, we average filter the image \(I^A\), \(I^B\) and \(I^C\) with an all-one matrix of M × M to obtain local information. 
\begin{equation}
	I^{C}_\mathcal{P}({p}_i)=\frac{1}{M * M}\sum_{p_j\in \mathcal{P}_i}I^C(p_j), {p_i\in \mathcal{R}_{12}},
\end{equation}
where \(\mathcal{P}_i\) is the image patch of M × M centered on \(p_i\). Similarly, we can get \(I^A_\mathcal{P}({p}_i)\) and \(I^B_\mathcal{P}({p}_i)\) for \({p_i\in \mathcal{R}_{12}}\). Within the seam area, \(I^C_\mathcal{P}({p}_i)\) has the information from \(I^A\) and \(I^B\) simultaneously, while \(I^A_\mathcal{P}({p}_i)\) only from \(I^A\) and \(I^B_\mathcal{P}({p}_i)\) only from \(I^B\). When \(I^C_\mathcal{P}({p}_i)\)  subtracts either \(I^A_\mathcal{P}({p}_i)\) or \(I^B_\mathcal{P}({p}_i)\), the result can reflect pixel difference within
the seam area between \(I^A\) and \(I^B\). The smaller the result, the more similar \(I^A\) and \(I^B\) are in the seam area, which means the higher the seam quality. So we can convert \(loss_{pixel}\) to:

\begin{equation}
	\begin{aligned}
		loss_{patch}=\frac{1}{N_{12}}\sum_{{p}_i\in \mathcal{R}_{12}}min(&|I^C_\mathcal{P}({p}_i)-I^A_\mathcal{P}({p}_i)|,\\&
		|I^C_\mathcal{P}({p}_i)-I^B_\mathcal{P}({p}_i)|).
	\end{aligned}	
\end{equation}

The final loss function is summarized as:
\begin{equation}
	loss=w_1*loss_{non}+w_2*loss_{patch},
\end{equation}
where the \(w_1\) and \(w_2\) represent the weights of \(loss_{non}\) and \(loss_{patch}\) respectively.

\section{Experimental}
\subsection{Dataset and Implement Details}
\textbf{Dataset} We validate the performance of the proposed network in a public image stitching dataset UDIS-D. The dataset is a real-world dataset proposed in UDIS \cite{nie2021unsupervised}, where 10440 image pairs are used for training and 1106 image pairs are used for testing. Unlike virtual datasets, almost all real-world data have parallax, so seam prediction is necessary to eliminate fusion ghosting. The original images in the dataset are unaligned, and we use UDIS-Net to register them to obtain the warped images. All the seam prediction experiments described in this paper are based on them.
\begin{figure*}[!t]
	\centering
	\includegraphics[width=0.92\textwidth]{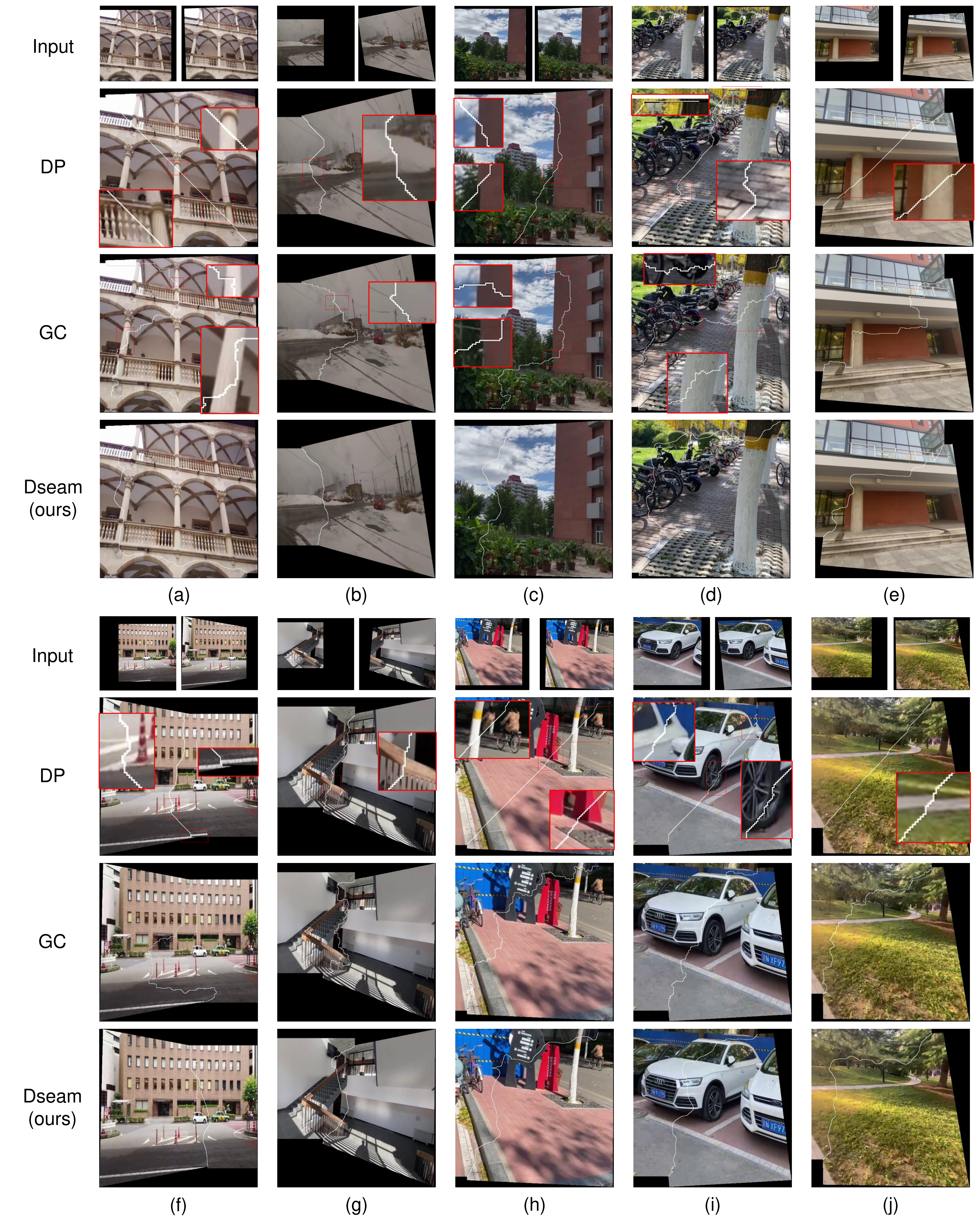} 
	\vspace{-3mm}
	\caption{Comparison of seam prediction results in various scenarios.}
	\vspace{-5mm}
	\label{visual}
\end{figure*}

\textbf{Implement Details} Our network is trained using an Adam optimizer \cite{kingma2014adam} with an exponentially decaying learning rate initialized to 0.0001 for 400k iterations. The batch size is set to 1. \(w_1\), \(w_2\) and M are assigned as 200, 100, and 9. We use RELU as the activation function for all the convolution layers except for the last layer. The implementation is based on TensorFlow and the network is performed on a single GPU with GeForce GTX 1080 Ti.

\subsection{Visual Comparison}

The visual performance of stitched image is a direct way to reflect the quality of seams. To compare different seam prediction methods, we apply dynamic programming (DP), GraphCut (GC) and our method to make seam prediction on the dataset UDIS-D. The codes for DP and GraphCut come from the seam prediction functions DpSeamFinder and GraphCutSeamFinder in OpenCV 2.4.9 \cite{bradski2000opencv}. The experimental results in Figure \ref{visual} show that: in a few cases, our results are better than both DP and GC, such as (a)(b)(c)(d); In most cases, our results are comparable to GC and better than DP, such as (e)(f)(g)(h)(i)(j).

To pursue high speed, DP reduces the requirements for seam quality, and only finds a local optimal solution by path search. Since DP only search the path in a fixed direction, its degree of freedom is too low to avoid some pixel misalignment areas. Especially when the start and end points are far apart, DP can only predict a seam that is approximately a straight line, such as (a)(e)(h)(j). In contrast, our method has high enough degree of freedom to avoid arbitrary areas and find a high-quality seam, while reaching a high speed.

GC is the most classical method of seam prediction, it transforms the seam prediction into a min-cut problem, and obtains the global optimal solution by energy optimization. Since its energy function is calculated based on the difference of individual pixels, the global optimal solution is only established in quantitative evaluation. In a few cases, the theoretically optimal seam doesn't lead to the best visual performance, such as (a)(b)(c)(d). In contrast, our method incorporates local information when calculating the loss, and pays more attention to local similarity. Although the location of seams predicted by our method is different from that of GC, the stitched images achieve similar visual performance to GC, and even better in a few cases.

\subsection{Seam Quality}
\subsubsection{Metrics} To evaluate the quality of seams, we use the metric proposed by SEAGULL \cite{lin2016seagull} and further used in \cite{li2018perception}. For each pixel \(p_i\) on the seam, they define a N × N local patch centered at \(p_i\). Then they calculate the ZNCC (zero normalized cross correlation) scores of the local patches between two images, and normalize the results. The seam quality is defined as follows:
\begin{equation}
	Q_{seam}(p) = \frac{1}{K}\sum_{i=1}^{K}(1.0 - \frac{ZNCC(p_i) + 1}{2}),
\end{equation}
where ZNCC is the local similarity score of the two images, and K is the number of pixels on the seam. The smaller \(Q_{seam}\) means the two images are more similar  near the seam and the predicted seam quality is higher.

In SEAGULL, the patch size (N) is set to 15. However, it is unfair to evaluate seams on a fixed scale, because the optimal scale for predicting seams may be different among different methods. Therefore, in order to get more robust results, we conduct experiments on the cases where N ranges from 2 to 15 to evaluate these methods on different scales.

\subsubsection{Analysis}  
\begin{figure}[t]
	\centering
	\includegraphics[width=1.0\columnwidth]{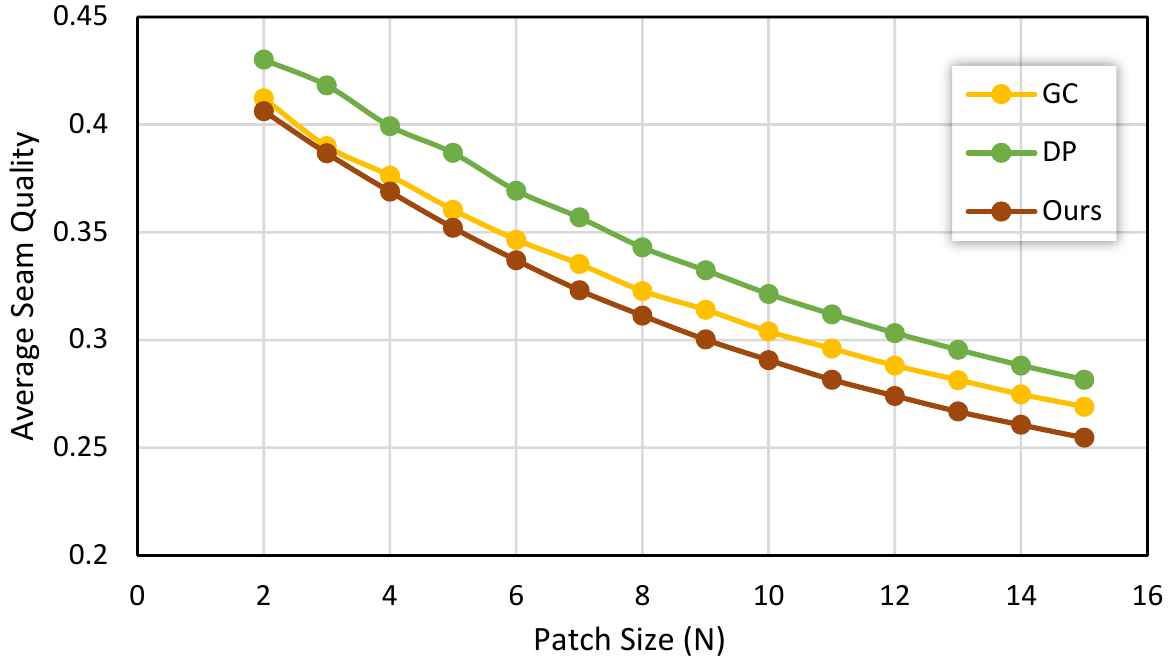} 
	\vspace{-7mm}
	\caption{Seam quality (lower is better) varies with N.}
	\vspace{-5mm}
	\label{seam_q}
\end{figure}

As shown in Figure \ref{seam_q}, the \(Q_{seam}\) of our method is always lower than that of DP and GC, which means that our method is better under the measure of Equation (8). The seam quality of DP is much worse than that of our method at any value of N. In contrast, our method is only slightly better than GC when N is small, but with the increase of N, the gap becomes more and more obvious.

According to Equation (8), the metric of seam quality is patch-based, while the cost functions of GC and DP are both pixel-based. For this reason, GC and DP perform better when patch size is small. And since the loss function incorporates local information, our method performs well at any scale. Although this metric cannot fully represent the quality of the seam, the experimental results still reflect the high quality of our seams to a certain extent.
\subsection{Speed}
	To make a fair comparison, all of our speed tests are performed on a server equipped with Intel Xeon CPU E5-2620 v4 @ 2.10GHz and GeForce GTX 1080 Ti. Both GC and DP are implemented using the official functions in Opencv2.4.9, and all results are generated with default parameters. We count the total time of seam prediction for the whole test dataset and calculate the average time and frame frequency. The statistical result of all methods include the time of image preprocessing, scaling for GC and DP, while scaling and edge extraction for our methods.
	\vspace{-3mm}
	\begin{table}[!h]
		\centering
		\resizebox{1\columnwidth}{!}{
			\vspace{-2mm}
			\begin{tabular}{c|c|c|c|c}
				\hline
				Method & Data & Time(s) & Average(s) & Frame(fps)\\
				\hline
				GC & 1106 & 95.5 & 0.0863 & 11.6\\
				Dp & 1106 & 7.6 & 0.0068 & 145.5\\
				Dseam(ours) & 1106 & 7.8 & 0.0071 & 141.8\\
				Dseam*(ours) & 1105 & 6.4 & 0.0058 & 172.7\\
				\hline
			\end{tabular}}
			\vspace{-1mm}
			\caption{Speed comparison of different methods.}
			\vspace{-3mm}
			\label{table1}
		\end{table}
	
	As shown in Table \ref{table1}, although we have downsampled the input images, GC can only reach a speed of 11.6 FPS due to its high complexity. In contrast, our method and DP both achieve high speeds (over 140 FPS), which are much faster than that of GC. In addition, due to network deployment, the deep learning method takes much more time to process the first frame than other frames. So we also only count the remaining frames except the first frame (Dseam* in Table 1). The result show that the actual speed of our method even exceed DP, about 15 times faster than GC.
	
	Seam prediction can be applied not only in image stitching, but also in video stitching which requires a high speed of algorithm. Since GC needs to iterate over and over again to optimize the energy function, it takes a long time and is not suitable for video stitching task. Thus, researchers can only use dynamic programming with poor quality but high speed to predict the seams in video stitching. However, our method combines the advantages of GC and DP, which can not only get high-quality seams, but also meet the speed requirements of video stitching.

	\subsection{Comparison with Non-seam Methods}
	\begin{figure}[h]
		\centering
		\vspace{-5mm}
		\includegraphics[width=0.95\columnwidth]{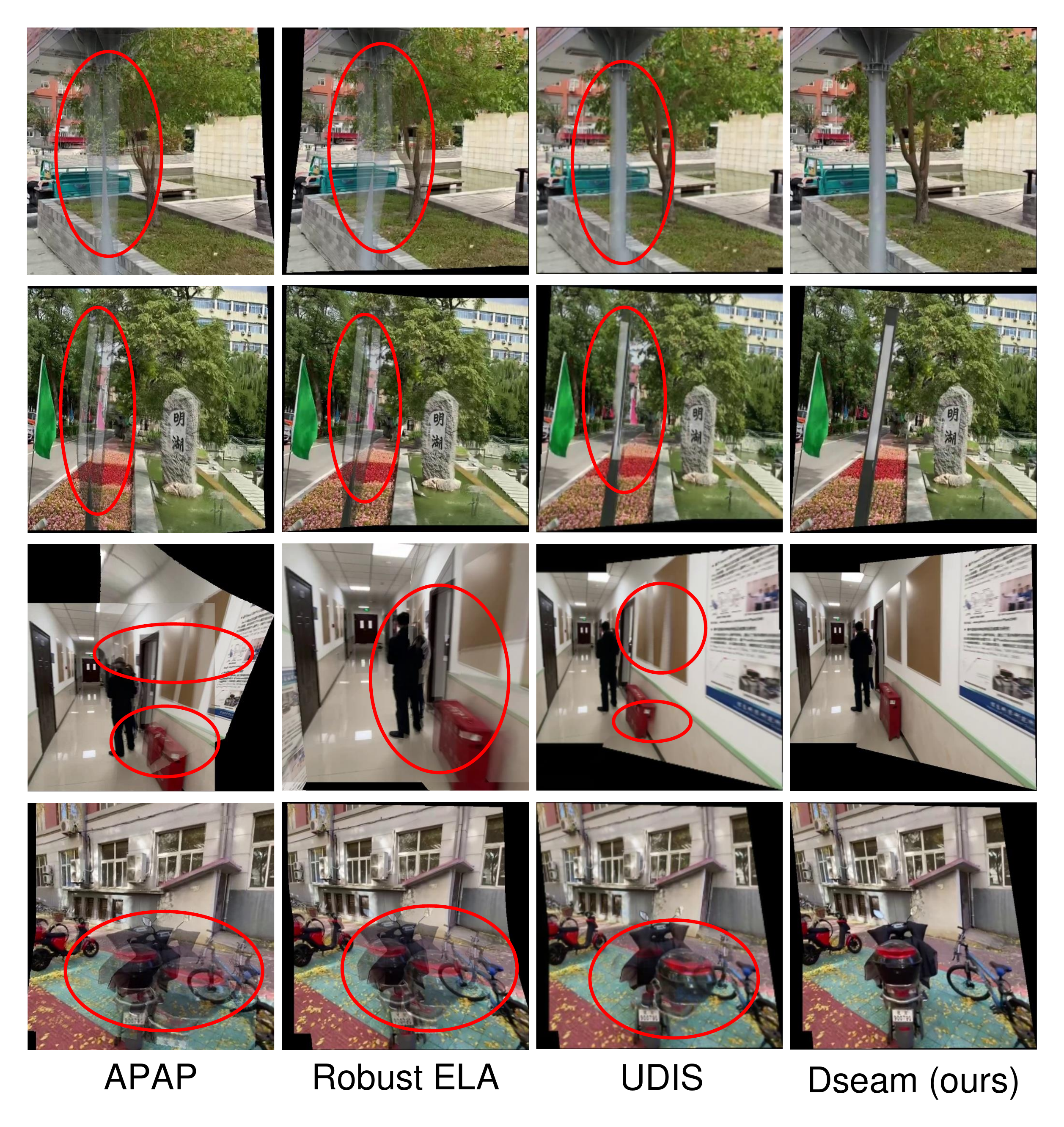} 
		\vspace{-5mm}
		\caption{Comparison with non-seam stitching methods.}
		\vspace{-3mm}
		\label{UDIS}
	\end{figure}
	For the parallax problem in image stitching, existing methods have tried to solve it from other view besides seam prediction. Therefore, we also compare with some of these non-seam stitching methods. Among these methods, APAP \cite{zaragoza2013projective} and Robust ELA \cite{li2017parallax} perform mesh alignment in the image registration stage, while UDIS \cite{nie2021unsupervised} fuses images in the feature domain. As shown in Figure \ref{UDIS}, for the case of large parallax, these  methods cannot solve the fusion ghosting problem well. In contrast, the stitched images of our method have good visual performance despite no subsequent image fusion operation.
	
	\subsection{Ablation Studies}
	In this section, we perform ablation experiments on our seam prediction method, which verify the effectiveness of our edge extraction operations on the input images and find the optimal size of the smoothing matrix. The seam quality assessment in these experiments is taken from the metric function at N=15. 
	\begin{figure}[h]
		\centering
		\vspace{-1mm}
		\includegraphics[width=1.0\columnwidth]{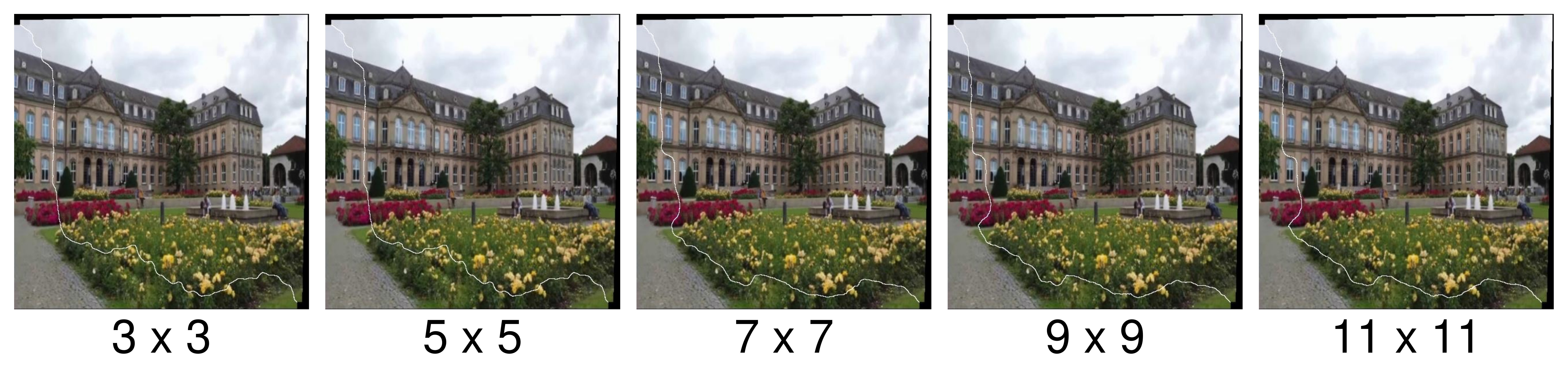} 
		\vspace{-7mm}
		\caption{Effect of matrix size on visual performance.}
		\vspace{-3mm}
		\label{size}
	\end{figure}
	
	As mentioned above, we need to use an all-one matrix of M × M to calculate the selection consistency loss. We carry out experiments on different values of the matrix size (M), and the experimental results are shown in Figure \ref{size} and Table \ref{table2}. As we can see, the seam locations are very similar for different matrix sizes, and the seam quality is also not much different. These demonstrate that our method is robust and insensitive to matrix size. In general, the seam quality of 9×9 is the highest, so we set M to 9 in the final version.
	\begin{table}[h]
		\centering
			\begin{tabular}{c|c|c}
				\hline
				Input Image &Matrix Size (M x M) & Seam Quality ($\downarrow$) \\
				\hline
				RGB &9 × 9& 0.259501 \\
				\hline
				Gray &9 × 9& 0.256870 \\
				\hline
				& 3 × 3 & 0.264690 \\
				&5 × 5 & 0.260572 \\
				\textbf {Edge}&7 × 7 & 0.255327 \\
				&\textbf{9 × 9} & \textbf{0.254621} \\
				&11 × 11 & 0.255517 \\
				\hline
			\end{tabular}
			\vspace{-2mm}
			\caption{Effect of input image and matrix size on quality.}
			\vspace{-3mm}
			\label{table2}
		\end{table}
		
		Since the seam prediction is sensitive to the edge, we use the edge image extracted by Sobel operator in both image input and loss calculation. In order to verify the effectiveness of edge extraction, we also conduct experiments on RGB image and gray image. As shown in Table \ref{table2}, the quality of seam predicted by edge image is better than that by RGB image and gray image.
		
		\section{Conclusion}
		In this paper, we propose a deep seam prediction method, which introduces deep learning into the field of seam prediction for the first time. In order to define the seam properly, we transform the seam prediction into mask prediction, and propose the selection consistency loss to make the network trainable. We adopt the classic encoding-decoding network as backbone, and use Sobel operator to extract the edge information which is more important for seam prediction, so that the network can achieve the best performance at limited network depth. The experimental results show that our method is superior to other advanced solutions and has the advantages of high quality and fast speed at the same time. Our work mainly focus on selecting consistency loss and pipeline design, but how to design a more appropriate network for this task is worth exploring in the future.

{\small
\bibliographystyle{ieee_fullname}
\bibliography{DSP}

\begin{thebibliography}{10}\itemsep=-1pt

\bibitem{bay2006surf}
Herbert Bay, Tinne Tuytelaars, and Luc~Van Gool.
\newblock Surf: Speeded up robust features.
\newblock In {\em European conference on computer vision}, pages 404--417.
  Springer, 2006.

\bibitem{bradski2000opencv}
Gary Bradski.
\newblock The opencv library.
\newblock {\em Dr. Dobb's Journal: Software Tools for the Professional
  Programmer}, 25(11):120--123, 2000.

\bibitem{burt1983multiresolution}
Peter~J Burt and Edward~H Adelson.
\newblock A multiresolution spline with application to image mosaics.
\newblock {\em ACM Transactions on Graphics (TOG)}, 2(4):217--236, 1983.

\bibitem{chalfoun2017mist}
Joe Chalfoun, Michael Majurski, Tim Blattner, Kiran Bhadriraju, Walid Keyrouz,
  Peter Bajcsy, and Mary Brady.
\newblock Mist: accurate and scalable microscopy image stitching tool with
  stage modeling and error minimization.
\newblock {\em Scientific reports}, 7(1):1--10, 2017.

\bibitem{duplaquet1998building}
Marie-Lise Duplaquet.
\newblock Building large image mosaics with invisible seam lines.
\newblock In {\em Visual information processing VII}, volume 3387, pages
  369--377. SPIE, 1998.

\bibitem{fischler1981random}
Martin~A Fischler and Robert~C Bolles.
\newblock Random sample consensus: a paradigm for model fitting with
  applications to image analysis and automated cartography.
\newblock {\em Communications of the ACM}, 24(6):381--395, 1981.

\bibitem{gaddam2016tiling}
Vamsidhar~Reddy Gaddam, Michael Riegler, Ragnhild Eg, Carsten Griwodz, and
  P{\aa}l Halvorsen.
\newblock Tiling in interactive panoramic video: Approaches and evaluation.
\newblock {\em IEEE Transactions on Multimedia}, 18(9):1819--1831, 2016.

\bibitem{gao2013seam}
Junhong Gao, Yu Li, Tat-Jun Chin, and Michael~S Brown.
\newblock Seam-driven image stitching.
\newblock In {\em Eurographics (Short Papers)}, pages 45--48, 2013.

\bibitem{jia2021leveraging}
Qi Jia, ZhengJun Li, Xin Fan, Haotian Zhao, Shiyu Teng, Xinchen Ye, and
  Longin~Jan Latecki.
\newblock Leveraging line-point consistence to preserve structures for wide
  parallax image stitching.
\newblock In {\em Proceedings of the IEEE/CVF conference on computer vision and
  pattern recognition}, pages 12186--12195, 2021.

\bibitem{kanopoulos1988design}
Nick Kanopoulos, Nagesh Vasanthavada, and Robert~L Baker.
\newblock Design of an image edge detection filter using the sobel operator.
\newblock {\em IEEE Journal of solid-state circuits}, 23(2):358--367, 1988.

\bibitem{kim2019deep}
Hak~Gu Kim, Heoun-Taek Lim, and Yong~Man Ro.
\newblock Deep virtual reality image quality assessment with human perception
  guider for omnidirectional image.
\newblock {\em IEEE Transactions on Circuits and Systems for Video Technology},
  30(4):917--928, 2019.

\bibitem{kingma2014adam}
Diederik~P Kingma and Jimmy Ba.
\newblock Adam: A method for stochastic optimization.
\newblock {\em arXiv preprint arXiv:1412.6980}, 2014.

\bibitem{kwatra2003graphcut}
Vivek Kwatra, Arno Sch{\"o}dl, Irfan Essa, Greg Turk, and Aaron Bobick.
\newblock Graphcut textures: Image and video synthesis using graph cuts.
\newblock {\em Acm transactions on graphics (tog)}, 22(3):277--286, 2003.

\bibitem{leutenegger2011brisk}
Stefan Leutenegger, Margarita Chli, and Roland~Y Siegwart.
\newblock Brisk: Binary robust invariant scalable keypoints.
\newblock In {\em 2011 International conference on computer vision}, pages
  2548--2555. Ieee, 2011.

\bibitem{li2017medical}
Desheng Li, Qian He, Chunli Liu, and Hongjie Yu.
\newblock Medical image stitching using parallel sift detection and
  transformation fitting by particle swarm optimization.
\newblock {\em Journal of Medical Imaging and Health Informatics},
  7(6):1139--1148, 2017.

\bibitem{li2017parallax}
Jing Li, Zhengming Wang, Shiming Lai, Yongping Zhai, and Maojun Zhang.
\newblock Parallax-tolerant image stitching based on robust elastic warping.
\newblock {\em IEEE Transactions on multimedia}, 20(7):1672--1687, 2017.

\bibitem{li2022automatic}
Jiaxue Li and Yicong Zhou.
\newblock Automatic color image stitching using quaternion rank-1 alignment.
\newblock In {\em Proceedings of the IEEE/CVF Conference on Computer Vision and
  Pattern Recognition}, pages 19720--19729, 2022.

\bibitem{li2018perception}
Nan Li, Tianli Liao, and Chao Wang.
\newblock Perception-based seam cutting for image stitching.
\newblock {\em Signal, Image and Video Processing}, 12(5):967--974, 2018.

\bibitem{li2015dual}
Shiwei Li, Lu Yuan, Jian Sun, and Long Quan.
\newblock Dual-feature warping-based motion model estimation.
\newblock In {\em Proceedings of the IEEE International Conference on Computer
  Vision}, pages 4283--4291, 2015.

\bibitem{liao2019quality}
Tianli Liao, Jing Chen, and Yifang Xu.
\newblock Quality evaluation-based iterative seam estimation for image
  stitching.
\newblock {\em Signal, Image and Video Processing}, 13(6):1199--1206, 2019.

\bibitem{lin2016seagull}
Kaimo Lin, Nianjuan Jiang, Loong-Fah Cheong, Minh Do, and Jiangbo Lu.
\newblock Seagull: Seam-guided local alignment for parallax-tolerant image
  stitching.
\newblock In {\em European conference on computer vision}, pages 370--385.
  Springer, 2016.

\bibitem{ng2003sift}
Pauline~C Ng and Steven Henikoff.
\newblock Sift: Predicting amino acid changes that affect protein function.
\newblock {\em Nucleic acids research}, 31(13):3812--3814, 2003.

\bibitem{nie2021depth}
Lang Nie, Chunyu Lin, Kang Liao, Shuaicheng Liu, and Yao Zhao.
\newblock Depth-aware multi-grid deep homography estimation with contextual
  correlation.
\newblock {\em arXiv preprint arXiv:2107.02524}, 2021.

\bibitem{nie2021unsupervised}
Lang Nie, Chunyu Lin, Kang Liao, Shuaicheng Liu, and Yao Zhao.
\newblock Unsupervised deep image stitching: Reconstructing stitched features
  to images.
\newblock {\em IEEE Transactions on Image Processing}, 30:6184--6197, 2021.

\bibitem{perez2003poisson}
Patrick P{\'e}rez, Michel Gangnet, and Andrew Blake.
\newblock Poisson image editing.
\newblock In {\em ACM SIGGRAPH 2003 Papers}, pages 313--318. 2003.

\bibitem{ronneberger2015u}
Olaf Ronneberger, Philipp Fischer, and Thomas Brox.
\newblock U-net: Convolutional networks for biomedical image segmentation.
\newblock In {\em International Conference on Medical image computing and
  computer-assisted intervention}, pages 234--241. Springer, 2015.

\bibitem{rublee2011orb}
Ethan Rublee, Vincent Rabaud, Kurt Konolige, and Gary Bradski.
\newblock Orb: An efficient alternative to sift or surf.
\newblock In {\em 2011 International conference on computer vision}, pages
  2564--2571. Ieee, 2011.

\bibitem{tian2019sosnet}
Yurun Tian, Xin Yu, Bin Fan, Fuchao Wu, Huub Heijnen, and Vassileios Balntas.
\newblock Sosnet: Second order similarity regularization for local descriptor
  learning.
\newblock In {\em Proceedings of the IEEE/CVF Conference on Computer Vision and
  Pattern Recognition}, pages 11016--11025, 2019.

\bibitem{wang2020multi}
Lang Wang, Wen Yu, and Bao Li.
\newblock Multi-scenes image stitching based on autonomous driving.
\newblock In {\em 2020 IEEE 4th Information Technology, Networking, Electronic
  and Automation Control Conference (ITNEC)}, volume~1, pages 694--698. IEEE,
  2020.

\bibitem{zaragoza2013projective}
Julio Zaragoza, Tat-Jun Chin, Michael~S Brown, and David Suter.
\newblock As-projective-as-possible image stitching with moving dlt.
\newblock In {\em Proceedings of the IEEE conference on computer vision and
  pattern recognition}, pages 2339--2346, 2013.

\bibitem{zhang2014parallax}
Fan Zhang and Feng Liu.
\newblock Parallax-tolerant image stitching.
\newblock In {\em Proceedings of the IEEE Conference on Computer Vision and
  Pattern Recognition}, pages 3262--3269, 2014.

\end{thebibliography}
}

\end{document}